\documentclass[10pt,twocolumn,letterpaper]{article}

\usepackage{cvpr}      %

\usepackage[dvipsnames]{xcolor}

\usepackage{colortbl}

\usepackage{bm}
\usepackage{bbm}

\usepackage{stfloats}

\usepackage{ifthen} 
\usepackage{multirow}
\usepackage{array}
\newcommand{\PreserveBackslash}[1]{\let\temp=\\#1\let\\=\temp}
\newcolumntype{C}[1]{>{\PreserveBackslash\centering}p{#1}}
\newcolumntype{L}[1]{>{\PreserveBackslash\raggedright}p{#1}}

\definecolor{lightgray}{gray}{0.8} %

\newcommand\blfootnote[1]{%
  \begingroup
  \renewcommand\thefootnote{}\footnote{#1}%
  \addtocounter{footnote}{-1}%
  \endgroup
}

\newcommand{\hide}[1]{}

\definecolor{Gray}{gray}{0.9}
\definecolor{LightGray}{gray}{0.95}

\DeclareMathOperator*{\argmax}{\arg\!\max}
\DeclareMathOperator*{\minover}{\min}

\newboolean{hidelegal}
\setboolean{hidelegal}{true}
\newcommand{\conditionaltext}[1]{%
    \ifthenelse{\boolean{hidelegal}}%
    {}%
    {#1}%
}

\newlength{\mysqueeze}
\usepackage[framemethod=TikZ]{mdframed}

\newmdenv[
  backgroundcolor=LightGray,
  linewidth=1pt,
  linecolor=Gray,
  roundcorner=7pt
]{myframe}

\definecolor{cvprblue}{rgb}{0.21,0.49,0.74}
\usepackage[pagebackref,breaklinks,colorlinks,citecolor=cvprblue]{hyperref}

\def\paperID{12630} %
\def\confName{CVPR}
\def\confYear{2024}

\title{Detours for Navigating Instructional Videos}

\usepackage{indentfirst}

\author{Kumar Ashutosh$^{1,2}$, Zihui Xue$^{1,2}$, Tushar Nagarajan$^{2}$, Kristen Grauman$^{1,2}$\\
$^{1}$UT Austin, $^{2}$FAIR, Meta\\
}

\begin{document}
\maketitle
\begin{abstract}
\blfootnote{Website: \href{https://vision.cs.utexas.edu/projects/detours}{https://vision.cs.utexas.edu/projects/detours}}
We introduce the \emph{video detours} problem for navigating instructional videos.  Given a source video and a natural language query asking to alter the how-to video's current path of execution in a certain way, the goal is to find a related ``detour video" that satisfies the requested alteration.  To address this challenge, we propose VidDetours, a novel video-language approach that learns to retrieve the targeted temporal segments from a large repository of how-to's using video-and-text conditioned queries.  Furthermore, we devise a language-based pipeline that exploits how-to video narration text to create weakly supervised training data.
We demonstrate our idea applied to the domain of how-to cooking videos, where a user can detour from their current recipe to find steps with alternate ingredients, tools, and techniques.  Validating on a ground truth annotated dataset of 16K samples, we show our model's significant improvements over best available methods for video retrieval and question answering, with recall rates exceeding the state of the art by 35\%.  %
\end{abstract}

\section{Introduction}
\label{sec:intro}

Instructional or ``how-to" videos are a compelling medium for people to share and learn new skills.  From everyday home fix-it projects, cooking, sports, to aspirational goals like playing piano beautifully, there are so many things that people of all ages and backgrounds want to learn or do a bit better.  Indeed, online how-to's are among the top few dominating categories of all content on YouTube, alongside entertainment and music. 
Advances in computer vision for keystep recognition~\cite{video-distant,task-graph-objective,graph2vid,task_graph,paprika,howto100m,mil-nce}, procedural task understanding~\cite{procedure-learning-fei-fei-li,procedure2,procedure3}, and video summarization~\cite{ivsum,vnd} have the potential to make such content more searchable and accessible.

\begin{figure}[t]
    \centering
    \includegraphics[width=0.47\textwidth]{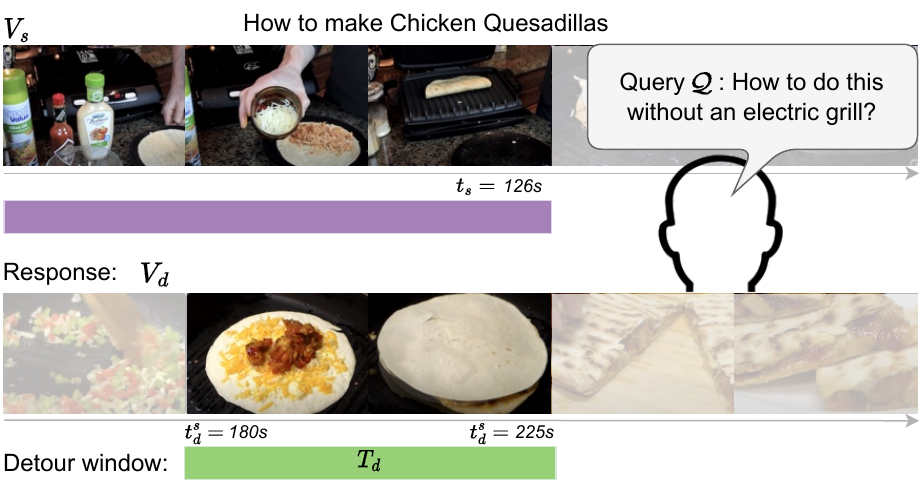}
    \caption{\textbf{An example video detour.} In the \emph{Chicken Quesadillas} recipe, the source video $V_s$ (top) shows the use of an electric grill at time instant $t_s$. A user watching this video does not have a grill and asks a query $\mathcal{Q}$ \emph{``how to do this without an electric grill?''}. In response, the system identifies a detour video $V_d$ and timepoint $T_d$ showing a similar recipe but using a heating pan instead of a grill.
    }
    \label{fig:teaser}
    \vspace{\mysqueeze} %
\end{figure}

However, while today's how-to content is a vast resource, it is
nonetheless disconnected. Human learners access how-to's in doses of one video at a time, studying the advice and visual demonstrations of one expert at a time. %
While there are thousands and thousands of videos addressing, for example,  \emph{``how to repair a bike'' or ``how to make a samosa"}, any given video offers only a single execution of a task using a fixed set of ingredients, tools, approach, and assuming a certain skill level.  When those criteria do not align, users face a dilemma whether to improvise, risking ``breaking'' the final output, or to find and watch another video hoping it better matches their constraints.  Manually synthesizing the information across videos is time consuming if not prohibitively taxing.

What if the wealth of knowledge in online instructional videos was not an array of isolated lessons, but instead an interconnected network of information?  What would it take to transform a pile of videos into a how-to knowledge base? 

Towards this vision, we explore how to intelligently navigate between related how-to videos,  %
conditioned on a natural language query.  Suppose a user watching a given video discovers they do not have the desired ingredients, tools, or skill-level. They may ask, \emph{``can I do this step without a wrench?''} or \emph{``I am on a diet, can I skip adding cheese here?''} or \emph{``how could I prepare the mix from scratch instead of using a pre-made one?''} or \emph{``is there a simpler way to do the corners?"} and so on.  Conditioned on the content watched so far in the \emph{source video}, the goal is to identify a \emph{detour video}---and a temporal segment within it---that would allow the user to continue their task with the adjustment specified by their language query, then return to the original source video and complete execution.
See Figure~\ref{fig:teaser}.

At the core this requires new technical advances in multimodal video understanding.  
Standard text-to-video retrieval models \cite{howto100m,videoclip,mil-nce,vid-retrieval-1,vid-retrieval-2,vid-retrieval-3,univl} are insufficient because the query text alone may not reveal enough details about the task (e.g., \emph{``can I do this step without an electric grill?''}) %
Similarly, existing video localization~\cite{2d-tan,vslnet,taco,umt} and question answering~\cite{naq,vqa-1,hero,tvqa+,just-ask,vqa-2} methods do not consider the viewing history of the source video, which is essential to properly identify a detour.  %
For example, answering \emph{``how to do this step without a wrench?''} requires the model to understand which steps are already done and which part of the target detour video shows the same effect \emph{without} using a wrench.

We introduce VidDetours: a video-language model that benchmarks this new problem.  Our approach formulates the video navigation task in two parts: (a)  retrieval of the detour video (b) temporal localization of the relevant portion of the detour video---both conditioned on the source video and text query.
Building on ideas from the video retrieval~\cite{mil-nce,videoclip,univl,youcook2} and localization~\cite{vslnet,2d-tan,stale} literature, we develop an architecture and training objective that accounts for all the essential components of this task: 
gauging the relatedness of any two instructional videos; capturing the \emph{interchangeability} of their component steps; and interactively indexing into the alternatives with language.

Since there are no existing datasets labeled for video detours, %
we devise a framework leveraging large language models (LLMs) to generate weakly-supervised training data.  Using HowTo100M \cite{howto100m}, a large-scale instructional video dataset of in-the-wild how-to's 
accompanied by the transcribed speech of the narrator, 
we automatically generate plausible user queries at 
targeted timepoints in training source videos together with their detour counterparts in closely related videos.  This procedure makes it possible to obtain ample effective training data without manual annotations.  To rigorously evaluate the video detours, we introduce a gold-standard test set comprised of manually labeled data from 4K full-length videos and 16K human-generated questions. 

In extensive experiments, we validate our model and illustrate the promise of the novel task.  VidDetours strongly outperforms state-of-the-art video-language and video retrieval methods.  We will release our training and test annotations to establish a formal benchmark for navigating instructional videos.  %

In short, ours is the first work to investigate personalized query-based navigation of instructional videos.  Our main contributions are the innovative task definition, our video-language model to address it, and the high quality eval set and benchmark.  These results help pave the way towards an interconnected how-to video knowledge base that would transcend the expertise of any one teacher, weaving together the myriad of steps, tips, and strategies available in existing large-scale video content.

\section{Related Work}
\label{sec:related}

\begin{figure*}[t]
    \centering
    \includegraphics[width=\textwidth]{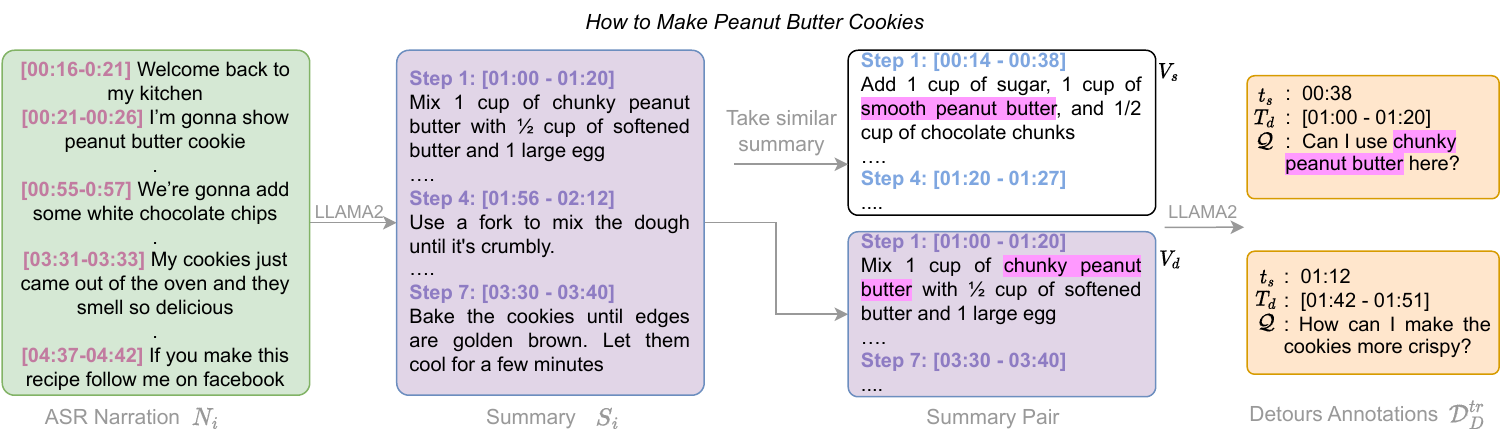}
    \caption{\textbf{Overview of the detours dataset ($\mathcal{D}_D^{tr}$) curation}. Given unlabeled instructional videos for training (we use HowTo100M~\cite{howto100m}), we first input their narrations with timestamps to a language model (LLAMA2~\cite{llama2}) to obtain summaries of their steps. %
    Next, we %
    automatically select pairs of similar summaries along with their timestamps and %
    use a language model to generate weakly-supervised detours annotation tuples $(V_s, t_s, \mathcal{Q}, V_d, T_d)$. As an example, the source video here uses smooth peanut butter.  A possible detour question is \emph{``can I use chunky peanut butter here?''} and the %
    window at $T_d$ in the detour video (top right, orange) shows the use of crunchy peanut butter.  %
    }
    \label{fig:summary-generation}
    \vspace{\mysqueeze} %
\end{figure*}

\indent \textbf{Learning from instructional videos.} 
Several recent video datasets like HowTo100M~\cite{howto100m}, COIN~\cite{coin}, CrossTask~\cite{crosstask} are based on instructional videos and have enabled research in procedure planning \cite{procedure-learning-fei-fei-li,procedure2,procedure3}, task graph learning \cite{task_graph,paprika,graph2vid,egoexo4d}, and alignment detection \cite{vnd,tan}. The availability of large-scale instructional videos on the internet also facilitates video representation learning %
for action recognition \cite{omnivore,mvitv2,uniformer,memvit,slowfast}, action anticipation \cite{avt,rulstm,intention,whenwillyoudowhat,gao2017red}, and object detection \cite{object-1,object-2}. All the prior work either focuses on short-term representations \cite{mil-nce,videoclip,univl,vast,valor} or video-level understanding \cite{task_graph,graph2vid}. To our knowledge, we are %
the first to establish a means to navigate across instructional videos, which is essential for holistic task understanding and optimal skill learning.

\textbf{Vision and language  learning.} Videos often also contain text---whether converted from narrations through automatic speech recognition (ASR)~\cite{howto100m,coin,crosstask} or manually annotated \cite{ego4d}. Using both text and video for representation learning \cite{mil-nce,videoclip,egovlp,hiervl,vast,valor} helps in multi-modal tasks like retrieval \cite{vid-retrieval-1,vid-retrieval-2,vid-retrieval-3,videoclip,univl}, localization \cite{naq,videoclip,taco,crosstask,actionformer,vslnet,umt,unloc}, captioning \cite{captioning-1,captioning-2,captioning-3,captioning-4,captioning-5}, question answering \cite{vqa-1,hero,tvqa+,just-ask,vqa-2}, and episodic memory queries~\cite{ego4d,egovlp,naq}.
Most of these tasks focus on images or clip-level understanding, typically a few seconds long. Recent work~\cite{hiervl,clip-hitchhiker,aggregation-1,aggregation-2,aggregation-3,ego4d,naq} extends this further for video-level understanding spanning minutes. None of the existing methods or benchmarks answer text queries by navigating \emph{between} %
long videos, as we propose.  %

\textbf{Interactive retrieval.} Dialog-based retrieval has been %
studied %
for fashion image retrieval~\cite{composed-fashion,ziad-fashion,fashion++} and conversation-based e-commerce shopping~\cite{ecommerce-1,ecommerce-2}, where a user wants a specific product and gives feedback on successive retrievals. In Visual Dialog~\cite{visual-dialog,visual-dialog-2,visual-dialog-3,visual-dialog-4,visual-dialog-5}, an agent is given an image and its caption and  has to answer questions about the image e.g. \emph{``what color is the mug?''} 
 Recent work in composed image/video retrieval \cite{covr,coir,coir-1,coir-2,coir-3,coir-4,whittle-search} uses an image/clip with a modification text to retrieve an improved version, e.g. a fountain image with text \emph{``at night''} retrieves a clip of the fountain at night. Similarly, StepDiff~\cite{stepdiff} %
 generates the difference between two clips in instructional videos. %

All the previous work focuses on improving video or image retrieval through dialog, and the inputs are image or short-duration video. %
In contrast, we focus on action demonstrations where the prompt can be about ingredients, tools, or even step executions, e.g., \emph{``how to prepare the mixture instead of using a pre-made mix?''} which is crucial for a holistic task-level understanding of many actions and dependencies amongst them. Furthermore, our setup considers full instructional videos that typically span several minutes, as opposed to static images and short clips. %

\section{Approach}
\label{sec:approach}
In this section, we first define our detour task formulation %
(Sec.~\ref{sec:setup}). Next, we detail the dataset collection process (Sec.~\ref{sec:dataset-gen}) and our model architecture (Sec. \ref{sec:method}). Finally, we discuss implementation and training details (Sec.~\ref{sec:implementation}).

\subsection{Video detour task formulation}
\label{sec:setup}

We define a \emph{video detour} as a mapping from a source video $V_{s}$ at timestamp $t_s$ to a response segment $T_d = (t_d^s, t_d^e)$ in a detour video $V_d$, based on a query text $\mathcal{Q}$. This is illustrated in Fig.~\ref{fig:teaser}, where after watching the source video for some time (purple bar, top panel), a user issues a query \emph{``how to do this without an electric grill?''}, for which the response is a segment in a different video showing the step in a pan instead of a grill (green bar, bottom panel).
By construction, $V_s$ and $V_d$ are related demonstrations from the same high-level task, that differ slightly in their demonstration (e.g., two videos demonstrating how to make chicken quesadillas).

Formally, we cast this as a video segment retrieval task conditioned on \emph{both a source video and a query text}. The goal is to find functions $\mathcal{F}_R$ and $\mathcal{F}_L$ such that
\begin{align}
V_d &= \argmax_{V_i \in \mathcal{D}} \mathcal{F}_R(V_i|V_s[1:t_s], \mathcal{Q}) \\
T_d &= \argmax_{T_i}  \mathcal{F}_L( T_i|V_s[1:t_s], \mathcal{Q}, V_d)     
\end{align}
where $V_s[1:t_s]$ refers to a video watched from the beginning, until time $t_s$, and $T_i$ refers to a temporal window (start and end time) in video $V_d$.

Here, $\mathcal{F}_R$ is a \emph{retrieval} mapping that finds the correct full instructional video, typically minutes long,
given the source video segment and the text query, while $\mathcal{F}_L$ is the \emph{localization} function that finds the start and end time in the detour video.

\subsection{Detour dataset generation}
\label{sec:dataset-gen}

Our goal is to learn retrieval and localization functions to find the correct detour segment given a source video and a query, for which we require a training dataset with tuples of the form $(V_s, t_s, \mathcal{Q}, V_d, T_d)$. The detour queries can focus on any aspect of the demonstration of the recipe: ingredients (e.g., \emph{``can I add eggs here?''}), tools, (e.g., \emph{``how to do this without an blender''}) or steps (e.g., \emph{``how do I serve this on a plate?''}). %
Existing procedural video datasets only offer a subset of the required information~\cite{howto100m,crosstask,coin,htstep-neurips2023}: they provide narrations or keystep labels, but are missing interrelations between different procedural demonstrations and thus cannot be used for detours training directly.
Moreover, collecting detour annotations at a large-scale may be impractical due to the amount of time required for annotators to watch and parse long detour videos. %

To address this, we propose an approach to automatically create a training dataset $\mathcal{D}_{D}^{tr}$ for our task using how-to video narrations and language models. Subsequently, for rigorous testing, we also manually collect ground truth test data $\mathcal{D}_{D}^{te}$ from human annotators. %

\textbf{Weakly-supervised training set $\mathcal{D}_{D}^{tr}$.} 
We start with a dataset of unlabeled instructional videos: $\mathcal{D} = \{(V_i, N_i)\}_{i=1}^{|\mathcal{D}|}$ containing \emph{narrations} $N_i$ in addition to the videos $V_i$. A narration is the spoken component of the how-to video, where the expert describes their actions (``now we mix for 3 minutes") and gives other commentary (``oh it looks great!").
 We concentrate on the broad domain of cooking due to its prominence in %
 instructional video datasets ($\sim$370K videos in HowTo100M~\cite{howto100m}), well-structured recipes, and strong interconnection between different instances. %

We use the narrations to generate labels for our training dataset $\mathcal{D}^{tr}_D$.
Despite being noisy, narrations have been used successively for weakly-supervised training labels for video-language pretraining \cite{howto100m,mil-nce,videoclip,univl} and keystep recognition \cite{task_graph,video-distant,paprika}. 
 Here we explore their utility for mining candidate detour pairs.  %
We do this in two stages.  %
See Figure~\ref{fig:summary-generation}.
First, we generate text summaries for the key steps in each video. %
Specifically, we prompt LLAMA 2 \cite{llama2}, a recent open source large language model, %
to summarize the instructional videos using the timestamped narrations $N_i$. %
The prompt is of the form ``Given the following narrations from a \conditionaltext{YouTube} video, what recipe is this, and %
summarize each step along with its timestamps...'' The exact text prompt %
is given in the Supp. along with example outputs. We obtain the summary of video $V_i$ as a tuple of the step start time, end time, and text description. %
This process yields an intermediate summary dataset; see Figure~\ref{fig:summary-generation} (second panel). %

Next, we generate detour queries and time windows given a pair of summaries. For this, we identify video pairs that share an activity (and therefore have similar summaries), as unrelated pairs are unlikely to yield a meaningful detour query $\mathcal{Q}$. Specifically, we sort summary pairs by cosine similarity of their MPNet~\cite{mpnet} sentence embeddings, discarding dissimilar pairs (score $<0.75$). With a video pair in hand, we design a prompt for the LLM
to generate detour queries and time windows $(t_s, \mathcal{Q}, T_d)$ of roughly the form ``Given video summaries with timestamps, suppose a person is watching video A, identify a text prompt that a user might issue to take a detour and watch video B, along with the detour timestamps? Some examples of detours ...''. See Supp. for full prompt details.
Note that a source video can have multiple valid queries and matching detour videos, which our generation strategy allows. %

\begin{figure}[t]
    \centering
    \includegraphics[width=0.45\textwidth]{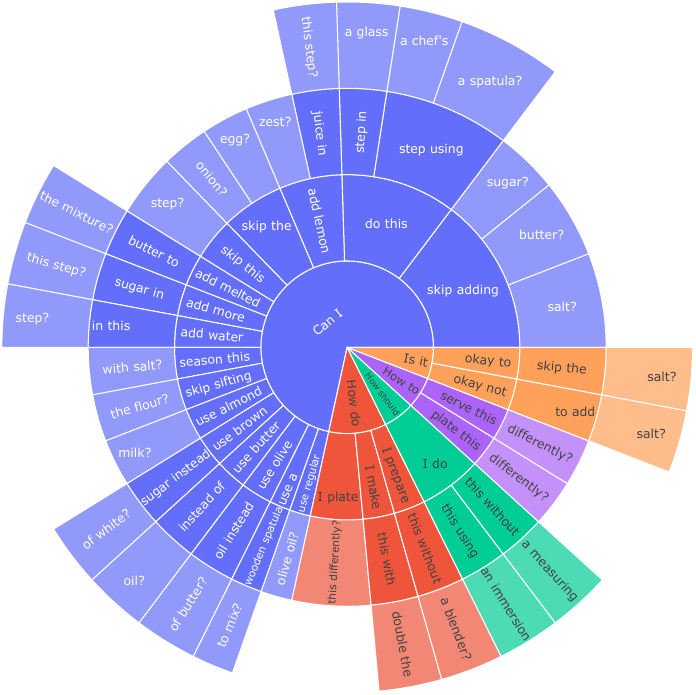}
    \caption{\textbf{Visualization of most frequent bigrams of the queries in the manually annotated test set}. We see that most of the queries have little or no context about the current recipe and the step being executed, e.g. \emph{``how do I plate this differently?''} or \emph{``can I do this step using a spatula?''}---emphasizing the need for source video context, as we explore in the proposed model.}
    \label{fig:sundial}
    \vspace{\mysqueeze} %
\end{figure}

The entire process yields a pair of videos and the detour annotations $(V_s, t_s, \mathcal{Q}, V_d, T_d)$ %
(Figure~\ref{fig:summary-generation}, right panel). More examples are in Supp. While %
$\mathcal{D}_D^{tr}$ will naturally have some noise due to language model errors and misaligned or non-visual narrations~\cite{tan,vnd}, we find them quite reasonable (85\% satisfactory) from manual inspection of a subset of the data, and our dataset creation strategy diminishes the noise.
Most importantly, they are effective for training a detour model, as our experiments testing on %
manually labeled data will show. %

\textbf{Manually collected testing set $\mathcal{D}_D^{te}$}. While the weakly-supervised data is sufficient for training, for reliable %
evaluation of our trained models and baselines, we manually collect ground truth test data.  Similar to $\mathcal{D}_D^{tr}$, we identify a pair of similar videos, and ask the professional annotators to watch the videos completely, and then annotate $(t_s, \mathcal{Q}, T_d)$. %
Annotators are allowed to reject pairs for which detours cannot be constructed (e.g., if they are too dissimilar). %
Since there can be multiple detours possible for a given pair of videos, we ask the annotators to identify at least three detours. %
We ensure that %
the videos in the train and test set are disjoint. The annotation process results in a high quality, benchmark test set for our new task.

Figure~\ref{fig:sundial} shows the diversity in queries that annotators provide. %
We see a variety of questions arise about substitutions of ingredients, tools, and steps. Further, we see contextual queries, e.g. \emph{``can I skip adding butter?''} that do not reveal much information about the recipe and the current step, underscoring the need to reference the source video when localizing a detour. %

\begin{figure*}[t]
    \centering
    \includegraphics[width=0.95\textwidth]{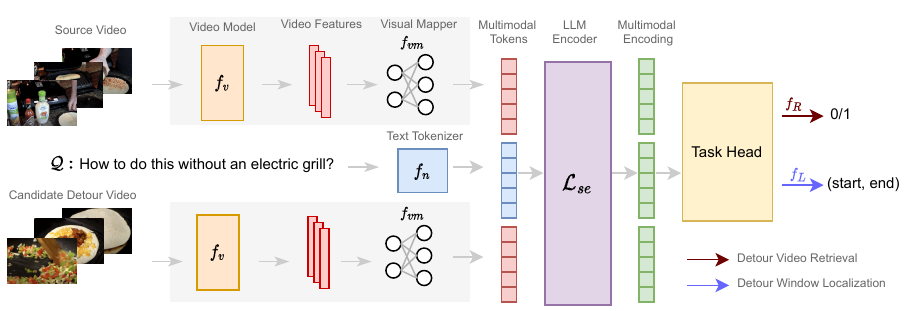}
    \caption{\textbf{Overview of the proposed approach}. The source video and the candidate detour video are converted to visual tokens by passing  through a video encoder $f_v$, followed by a visual mapper $f_{vm}$. We obtain a similar text token using standard tokenizer $f_n$. The processed tokens are then passed through a multimodal sequence encoder $\mathcal{L}_{se}$ to obtain output features. Finally, we have specific task heads for detour video retrieval and detour window localization. %
    }
    \label{fig:model}
    \vspace{\mysqueeze} %
\end{figure*}

\subsection{Detour retrieval and localization modules}
\label{sec:method}

Next, we describe our training framework to learn the mapping functions $\mathcal{F}_R$ and $\mathcal{F}_L$ using our generated detour dataset. %
Our objective is to design a multimodal (video and text) architecture that fuses the source video context with the language query enabling $\mathcal{F}_R$ and $\mathcal{F}_L$ to utilize both the viewing history and the query. As we will see in the experiments, this idea is more effective than late fusion of video and query features \cite{internvideo,clip,clip-hitchhiker}. 
For this, we leverage the reasoning capabilities of large language models~(LLM). In short, we encode both videos and the detour query as a sequence of tokens to pass to the LLM, which aggregates multimodal information from the inputs in its output encoding. These encodings are finally used to retrieve detour videos and detour segments using task specific heads. Our overall framework is shown in Fig.~\ref{fig:model} and each component is described in detail below.

\noindent \textbf{Tokenizing %
videos and detour queries.} %
We begin by encoding the videos as a sequence of tokens compatible with the LLM. We encode each video as spatio-temporal features using a video encoder $f_v$ (e.g., InternVideo~\cite{internvideo}), extracted at one feature per second~\cite{videoclip,task_graph} %
Next, we use a low-parameter visual mapper~$f_{vm}$~\cite{llava,llava-med,llama-adapter,video-chatgpt,video-llama} to convert the video features into visual tokens %
\textbf{v}~$=~f_{vm}(f_v(V))$ that are in the same embedding space as the text tokens, making them compatible with language encoders. %
Finally, we encode the detour query into text tokens using a 
standard text tokenizer (e.g. \cite{llama2,bert}) \textbf{n}$_{\mathcal{Q}} = f_n(\mathcal{Q})$. We now have both visual and query tokens in the same space to feed into our language models. %

\noindent \textbf{LLMs as multimodal sequence encoders.} Next, we use a LLAMA2 model as our multimodal sequence encoder ($\mathcal{L}_{se})$. LLMs are an ideal choice here, given our goal of encoding \emph{dialogue-driven detours}---the idea that a user is watching the source video and pauses it to ask a query, followed by a response from the LLM. %
Specifically, we
append source video tokens (until time $t_s$) with the query tokens $\text{\textbf{v}}_s[1:t_s] ~ | ~ \text{\textbf{n}}_{\mathcal{Q}}$, followed by candidate response video tokens. The multimodal encoder captures the cross-modal interactions between the source video, query context and the candidate video, resulting in an encoded output for the candidate video $O_i  = \mathcal{L}_{se} \left(   \text{\textbf{v}}_s[1:t_s] ~ | ~ \text{\textbf{n}}_{\mathcal{Q}} ~ | ~ \text{\textbf{v}}_i  \right)$. %

\noindent \textbf{Detour retrieval and localization heads.} Finally, we use the updated multimodal encodings $O_i$ to score detour video candidates and segments using two task heads $f_R$ and $f_L$. %
$f_R$ is a classifier that scores the relevancy of candidate video $V_i$ given the viewing context and the user query, while  %
$f_L$ is a classifier that identifies the highest score time segment $T_i$ inside the correct detour video $V_d$, given $O_d$. These two heads are trained separately for each task.

For our video retrieval network $\mathcal{F}_R$, we minimize the binary cross entropy loss $f_{BCE}$ between the prediction and the ground truth label: 
\begin{align*}
    \minover_{V_i \in \mathcal{D}} f_{BCE}(\mathcal{F}_R(V_i | V_s[1:t_s], \mathcal{Q}), \mathbbm{1}(V_i, V_d)).
\end{align*}

We assign a positive label to the correct detour video $V_d$ and negative to other videos sampled from the dataset, i.e. $V_i \neq V_d$. In particular, for every correct training instance, we randomly sample %
an incorrect video from which to curate a negative sample---either from the same task (hard negatives) or other tasks. 

The localization network $\mathcal{F}_L$ training objective is:
\begin{align*}
    \minover_{T_i = [t_i^s, t_i^e]} \frac{1}{2} \left[ f_{CE}(t_i^s, t_d^s) + f_{CE}(t_i^e, t_d^e) \right],
\end{align*}
where $f_{CE}$ is the cross-entropy loss. %
Similar to the video-language grounding model VSLNet~\cite{vslnet}, this objective minimizes the error in the distribution of the start and end times across the video. %
At inference, we find the candidate video $V_d$ and time duration $T_d$ that maximizes the scores from  %
$\mathcal{F}_R$ and $\mathcal{F}_L$, respectively (Sec.~\ref{sec:setup}). Note that there may be multiple plausible detour videos for a given source and query (e.g., multiple videos can use a \emph{heating pan} instead of an \emph{electric grill}); %
our scoring-based approach ensures other related (and valid) pairings can also score highly.

\begin{table*}[t]\footnotesize
    \centering
    \setlength{\tabcolsep}{0pt} %
    \begin{tabular}{L{3.2cm}|C{1.1cm}C{1.1cm}C{1.1cm}C{1.1cm}}
    \hline
        Method & R@5 & R@10 & R@50 & MedR $\downarrow$   \\
        \hline
        \arrayrulecolor{lightgray} %
        \textcolor{gray}{Text-only}  & \textcolor{gray}{3.9} & \textcolor{gray}{8.7} & \textcolor{gray}{14.0} & \textcolor{gray}{512} \\
        CLIP \cite{clip} & 7.9 & 11.8 & 25.2 & 342 \\
        CLIP-Hitchhiker \cite{clip-hitchhiker}  & 8.4 & 12.3 & 25.6 & 336 \\
        InternVideo \cite{internvideo} & 9.7 & 13.2 & 27.2 & 313 \\
        Distant Supervision \cite{video-distant} & 8.4 & 12.6 & 25.1 & 329 \\
        Multi-modal LLM \cite{vid2seq} & 5.9 & 10.5 & 32.1 & 139 \\
        CoVR \cite{covr}  & 4.3 & 9.2 & 15.3 & 473 \\
        \hline
        \rowcolor{Gray}
        Ours  & \textbf{17.6} & \textbf{27.8} & \textbf{62.4} & \textbf{30} \\
        \hline
        Ours w/o hard-negatives & 16.5 & 24.9 & 56.3 & 55 \\
        Ours w/ parser  & 13.9 & 21.6 & 50.0 & 81 \\
        \arrayrulecolor{black} %
        \hline
    \end{tabular}
    \hspace{1cm}
    \begin{tabular}{L{3.1cm}|C{1.2cm}C{1.2cm}C{1.2cm}C{1.2cm}}
    \hline
        Method & R@1, IoU=0.3  & R@1, IoU=0.5  & R@1, IoU=0.7 & Mean R@1   \\
        \hline
        \arrayrulecolor{lightgray} %
        \textcolor{gray}{Text-only} & \textcolor{gray}{5.2} & \textcolor{gray}{2.7} & \textcolor{gray}{0.6} & \textcolor{gray}{4.2} \\
        2D-TAN \cite{2d-tan}   & 10.3 & 4.2 & 1.5 & 8.6 \\
        VSLNet \cite{vslnet}   & 11.8 & 5.8 & 1.7 & 9.4 \\
        UMT \cite{umt}   & 12.0 & 6.1 & 1.6 & 9.4 \\
        Distant Supervision \cite{video-distant} & 10.6 & 4.0 & 1.5 & 8.3 \\
        Multi-modal LLM \cite{vid2seq} & 12.7 & 6.5 & 1.8 & 10.2 \\
        STALE \cite{stale}   & 12.1 & 6.1 & 1.7 & 9.6 \\
        \rowcolor{Gray}
        Ours   & \textbf{16.7} & \textbf{7.7} & \textbf{2.8} & \textbf{12.8} \\
        \hline
        Ours w/ parser   & 13.4 & 7.0 & 2.5 &  11.6 \\
        \arrayrulecolor{black} %
        \hline
    \end{tabular}
    
    \caption{Results for detour video retrieval (left) and detour window localization (right) tasks. Our method outperforms all prior methods and baselines by a significant margin.}
    \label{tab:all-results-main}
    \vspace{\mysqueeze} %
\end{table*}

\subsection{Implementation details}
\label{sec:implementation}

\textbf{Dataset and statistics.} Our training and test sets are both derived from HowTo100M \cite{howto100m} (average length 6.5 mins). We consider cooking tasks (i.e., recipes) containing $370K$ videos. We use the weakly-supervised automatically curated dataset $\mathcal{D}_D^{tr}$ for training and validation. Following the steps described in Sec. \ref{sec:dataset-gen}, we obtain $586,603$ training and $18,308$ validation detour annotation tuples $(V_s, t_s, \mathcal{Q}, V_d, T_d)$. %

The manual annotation results in a large-scale test dataset $\mathcal{D}_D^{te}$ containing $16,207$ detour instances $(V_s, t_s, \mathcal{Q}, V_d, T_d)$. The test set is based on $3,873$ unique videos across $1,080$ recipes/tasks, e.g. \emph{``how to make chicken quesadillas''} is one task and has multiple video instances. We curate the test set so that there are $834$ \emph{common} tasks ($14,450/16,207$ annotations) with the training data $\mathcal{D}_D^{tr}$ having videos from those tasks.   There are $246$ additional \emph{novel} recipes ($1,757/16,307$ annotations) that do not %
appear in %
the training set. For retrieval evaluation, the detour candidates are all videos in the dataset, i.e. $3,873$ candidates per detour annotation. No video exists in both the training and testing split. %

\textbf{Network architecture.} We use InternVideo \cite{internvideo} as the video feature extractor $f_v$. The features are extracted at one feature per second, following \cite{videoclip,tan}. $f_v$ is frozen and $f_{vm}$ is a trainable linear layer, inspired by \cite{llava}. $\mathcal{L}_{se}$ is a LLAMA2-13B-chat \cite{llama2} language model. We try both variants of keeping $\mathcal{L}_{se}$ frozen and trainable; performance is better if $\mathcal{L}_{se}$ is trainable. Lastly, $f_n$ is the LLAMA-2 tokenizer.
The retrieval head $f_R$ is a transformer classifier and we take the CLS token of the output followed by a linear layer to output the score. Finally, the localization head $f_L$ is the VSLNet \cite{vslnet} architecture. We remove the tokenizer from VSLNet since the text input is already processed.

\textbf{Training parameters.} We train both networks on 8 nodes with 8 NVIDIA A100 GPUs for 5 epochs. The training time is 8 hours. We use AdamW \cite{adamw} optimizer with learning rate $3\times 10^{-5}$ and batch size of 16 per device. The transformer classifier $f_R$ uses an input dimension of 4096 (consistent with the LLAMA2 output dimension), 4 heads, 4 layers and 1024 dimensional feed-forward network. All other parameters are defaults of the respective models.

\section{Experiments}
\label{sec:expts}

\begin{figure*}[t]
    \centering
    \includegraphics[width=\textwidth]{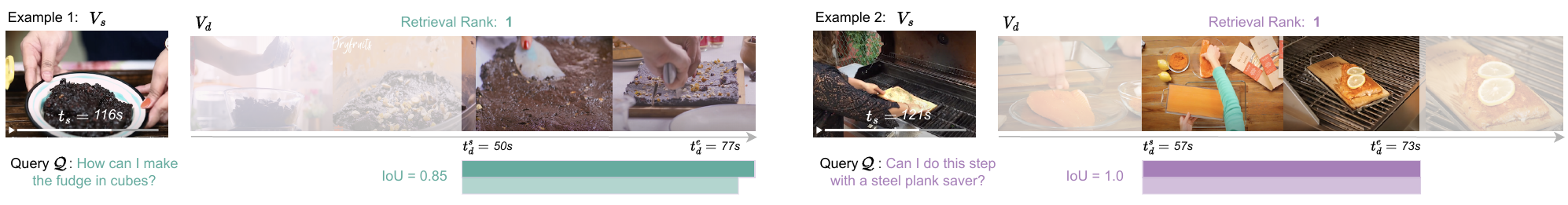} \\
    \includegraphics[width=\textwidth]{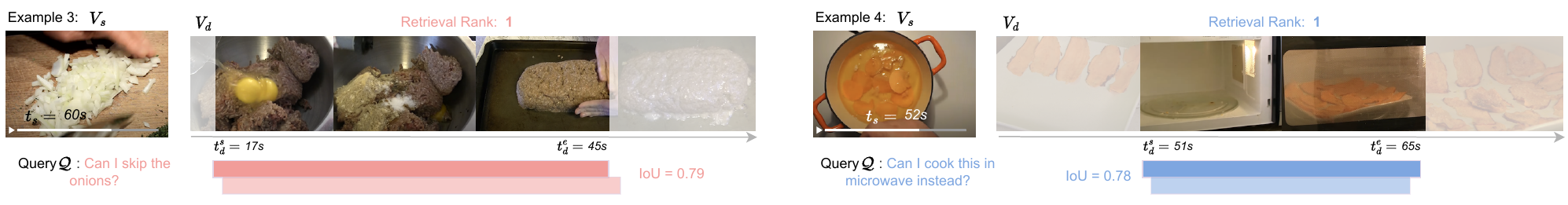} \\
    \includegraphics[width=\textwidth]{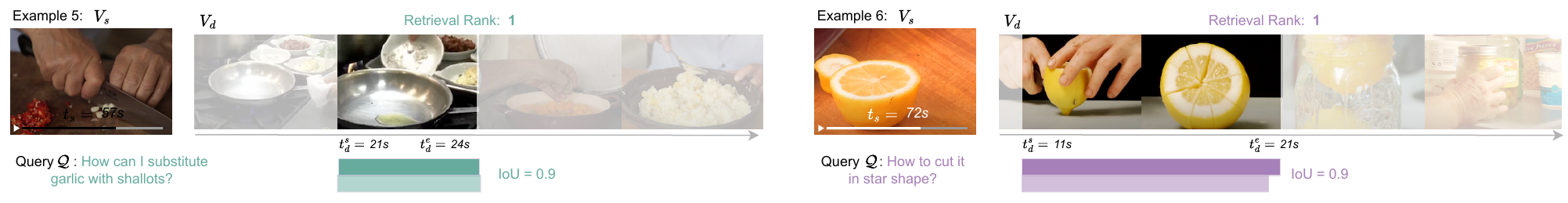} \\
    \vspace{-8pt}
    \color{gray}\rule{\linewidth}{0.1mm}
    \vspace{8pt}
    \includegraphics[width=\textwidth]{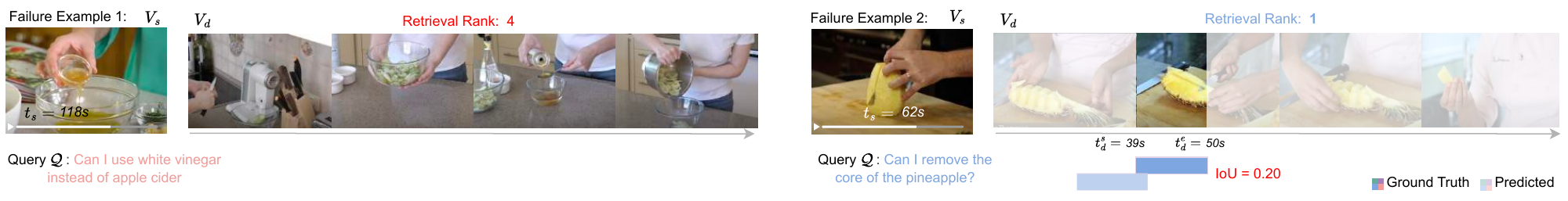}
    \vspace*{-0.3in}
    \caption{\textbf{Visualization of our model's predictions} for the detour video retrieval and window localization tasks, including failure cases (last row). For all the successful examples, our method ranks the detour video at top-1. Furthermore, in the detour window localization, our method is able to predict the detour window with a high overlap ($>0.7$ IoU). Our model is able to correctly use source video context and nuanced queries like \emph{``how to cut it in star shape?''} (Example 6) and \emph{``can I skip the onions?''} (Example 3). Best viewed with zoom. %
 }
    \label{fig:main_viz}
    \vspace{\mysqueeze} %
\end{figure*}

We show the results for the detour video retrieval (Sec. \ref{sec:retrieval} and detour localization (Sec. \ref{sec:localization}) subtasks.

\subsection{Detour video retrieval}
\label{sec:retrieval}

First, we benchmark models on finding the
correct detour video given the source video and the query, from amongst all videos in the test set (3,873 videos). %

\textbf{Baselines.} We adapt state-of-the-art video retrieval methods for detour retrieval. All the baselines embed text and video in a shared space using an encoder $\phi$. We find the detour video that has the most similar embedding to a reference $V_d = \argmax_{V_i \in \mathcal{D}} \left\langle \phi(V_i), \psi(V_s, \mathcal{Q})\right\rangle$, where $\psi(V_s, \mathcal{Q})$ computes the reference embedding from the source and query videos, and $\langle \rangle$ is cosine-similarity. We evaluate three variants corresponding to different inputs:
    \begin{itemize}
    \item \emph{with $V_s, \mathcal{Q}$}: where $\psi(V_s, \mathcal{Q}) = 1/2[\phi(V_s) + \phi(\mathcal{Q})]$.
    \item \emph{with $V_s$}: where $\psi(V_s, \mathcal{Q}) = \phi(V_s)$.
    \item \emph{with $\mathcal{Q}$}: where $\psi(V_s, \mathcal{Q}) = \phi(\mathcal{Q})$.
    \end{itemize}
    These variants test whether individual embeddings (or simple combinations of them) are sufficient for detour retrieval.

\begin{itemize}[leftmargin=*]
\itemsep0em 
    \item \textbf{Text-only} computes the similarity between summary and query embeddings, ignoring visual cues. This baseline evaluates the impact of %
    text-bias in the automatically curated dataset. %
    \item \textbf{CLIP \cite{clip}, InternVideo \cite{internvideo}} are state-of-the-art vision-language models used extensively for multi-modal tasks. Since they take short video clips as input, the video representation is the average of all short-term features.

    \item \textbf{CLIP-Hitchhiker \cite{clip-hitchhiker}} is similar to CLIP~\cite{clip} but uses a weighted average of frame features, instead of uniformly averaging for $\phi(V_i)$. %
    The weights are the similarity score between the frame and query features (eq. 1 in \cite{clip-hitchhiker}).
    Note that the \emph{source-only} variant is not evaluated as query features are needed to compute the weights.    
    \item \textbf{Distant Supervision \cite{video-distant}} uses WikiHow as an external knowledge base to map steps in the video with keysteps. %
    We replace the narrations with this keystep assignment for the detour dataset generation.
    \item \textbf{Multi-modal LLM \cite{vid2seq}} can be used to generate dense captions to replace the narrations. We use Vid2Seq \cite{vid2seq} to first densely annotate the videos and convert the task to text-only video detours using those captions.
    
    \item \textbf{CoVR \cite{covr}} is a state-of-the-art method for composed video retrieval. Following CoVR, we sample a frame from the source video and use the user query as the modification text to obtain detour retrieval. %
\end{itemize}

Note that we train \cite{video-distant, vid2seq, covr} on the same detour dataset as our model, whereas \cite{internvideo,clip,clip-hitchhiker} are used in zero-shot setting for its good retrieval capabilities in instructional videos.

\textbf{Ablations.} Similar to how we evaluate the baselines with different inputs, we compare the performance of our method (i.e. ``with $V_s, \mathcal{Q}$'') with ablations---``with $V_s$'' and ``with $\mathcal{Q}$''.
Next, recall that we use a language model to generate timestamps for both the summaries and the detour timestamps (see Fig. \ref{fig:summary-generation} and Sec. \ref{sec:dataset-gen}), which can be noisy. Instead, for this ablation, we parse the summaries and queries and assign timestamps based on the similarity score between the narrations and the summaries. The language model does not handle timestamps for this ``Ours w/ parser'' baseline. Finally, we evaluate the retrieval performance without sampling negatives from the same task, i.e. no hard negatives in ``Ours w/o hard negatives'' baseline.

\textbf{Metrics.} Following standard retrieval evaluation \cite{howto100m,mil-nce,videoclip,vina}, we report recall@$k$ for $k \in [5, 10, 50]$ and the median rank. 
Recall@$k$ ranges in $[0, 1]$, higher the better. The median rank ranges between $1$ and the size of the retrieval candidate set ($3,873$), the lower the better. %

\textbf{Results.} Table \ref{tab:all-results-main} (left) reports the results with both inputs, i.e. ``with $V_s, \mathcal{Q}$'' for all baselines on all metrics. We also report medR for different inputs in Table \ref{tab:retrieval-ablation} (left).
Our method outperforms all the baselines by a significant margin. Our performance gain is 7.9\% w.r.t. InternVideo \cite{internvideo} at R@5 and the gap further increases to \textbf{35.2\%} for R@50. Across all the different input combinations, InternVideo \cite{internvideo} ``with $\mathcal{Q}$'' is the second best and attains a medR of 138 --- much lower than our medR of 30 (Tab. \ref{tab:retrieval-ablation}).

We attribute our large gain to three crucial factors: (a) thanks to its tokens and long sequence length in the LLM, our model is capable of long video understanding, unlike CoVR \cite{covr}, which samples a few frames from the video; (b) appropriately using the source video context $V_d$, unlike CLIP \cite{clip}, CLIP-Hitchhiker \cite{clip-hitchhiker}, and InternVideo \cite{internvideo}, since our model is more than a standard text-to-video retrieval model; and (c) conditioning the retrieval on the query text. Furthermore, we find that using narrations for generating a weakly-supervised training set is better than using keysteps \cite{video-distant} or captions \cite{vid2seq}, %
which miss %
details typically present in narrations. %

Fig.~\ref{fig:main_viz} shows example detours inferred by our model.  
In all the successful cases, our retrieval model ranks the correct detour video at top-1. It is evident that detour video retrieval is more challenging than conventional text-to-video-retrieval because the detour query can have missing context, e.g. \emph{``how do I add garnish after pouring?''} does not reveal the task in the video; the previous video context is needed for the correct retrieval. The failure cases (bottom row) show example errors in retrieval rank (left) and localization (right). %
The performance trend is similar for \emph{common} and \emph{novel} tasks of the dataset (see Supp.), 
reinforcing %
that training on a sufficiently large dataset with many recipes enables good generalization. %

\subsection{Detour window localization}
\label{sec:localization}

\begin{table}\footnotesize
    \centering
    \setlength{\tabcolsep}{0pt} %
    \begin{tabular}{L{2.0cm}|C{0.5cm}C{0.5cm}C{1.1cm}}
    \hline
        Method & $V_s$ & $\mathcal{Q}$ & MedR $\downarrow$   \\
        \hline
        \arrayrulecolor{lightgray} %
        CLIP \cite{clip} &  \checkmark &  & 314 \\
        &   & \checkmark & 191 \\
        & \checkmark & \checkmark & 342 \\
         \hline
        CLIP-Hitch. \cite{clip-hitchhiker}  & \checkmark &  & --- \\
        &  & \checkmark & 186 \\
         & \checkmark & \checkmark & 336 \\
        \hline
        InternVideo \cite{internvideo} &  \checkmark &  & 150 \\
        &   & \checkmark & 138 \\
        & \checkmark & \checkmark & 313 \\
        \hline
        DistantSup. \cite{video-distant} &  \checkmark &  & 384 \\
        &   & \checkmark & 370 \\
        & \checkmark & \checkmark & 329 \\
        \hline
        MLLM \cite{vid2seq} &  \checkmark &  & 189 \\
        &   & \checkmark & 158 \\
        & \checkmark & \checkmark & 139 \\
        \hline
        CoVR \cite{covr}  & \checkmark &  & 388 \\
           &  & \checkmark & 401  \\
          & \checkmark & \checkmark & 473  \\
        \hline
        Ours  & \checkmark &  & 128  \\
        &  & \checkmark & 116  \\
        \rowcolor{Gray}
        & \checkmark & \checkmark & \textbf{30}  \\
        \arrayrulecolor{black} %
        \hline
    \end{tabular}
    \hspace{0.0cm}
        \begin{tabular}{L{1.9cm}|C{0.5cm}C{0.5cm}C{1.1cm}}
    \hline
        Method & $V_s$ & $\mathcal{Q}$ & R@1   \\
        \hline
        \arrayrulecolor{lightgray} %
        2D-TAN \cite{2d-tan} & \checkmark & & 5.5 \\
        &   & \checkmark & 8.0 \\
        & \checkmark  & \checkmark & 8.6 \\
         \hline
        VSLNet \cite{vslnet}  & \checkmark  & & 6.1 \\
        &   & \checkmark & 8.5 \\
         & \checkmark & \checkmark & 9.4 \\
        \hline
        UMT \cite{umt} & \checkmark & & 6.5 \\
        &   & \checkmark & 8.7 \\
        & \checkmark  & \checkmark & 9.4 \\
        \hline
        DistantSup. \cite{video-distant} &  \checkmark &  & 7.6 \\
        &   & \checkmark & 7.9 \\
        & \checkmark & \checkmark & 8.3 \\
        \hline
        MLLM \cite{vid2seq} &  \checkmark &  & 9.1 \\
        &   & \checkmark & 9.7 \\
        & \checkmark & \checkmark & 10.2 \\
        \hline
        STALE \cite{stale}   & \checkmark  & & 6.9 \\
           &   & \checkmark & 8.8  \\
          &  \checkmark & \checkmark & 9.6  \\
        \hline
        Ours  & \checkmark & & 8.9  \\
        &  & \checkmark &  11.2  \\
        \rowcolor{Gray}
        & \checkmark & \checkmark & \textbf{12.8}  \\
        \arrayrulecolor{black} %
        \hline
    \end{tabular}
    \caption{Comparison of our method with prior methods at different input combinations for detour video retrieval (left) and detour window localization (right). Our method outperforms all the prior works for all input combinations. See Supp. for all metrics.}
    \label{tab:retrieval-ablation}
\end{table}

Next, we show results on detours window localization where the task is to determine the correct window given the query, the source video, and the ground-truth detour video.  %

\textbf{Baselines.} While there are no existing methods %
that use previous minutes-long viewing history along with the query to perform temporal localization, we use state-of-the-art localization %
methods as baselines and also strengthen them by providing source video context, as described below.

\begin{itemize}
    \item \textbf{Text-only:} Similar to the detours video retrieval, we have a text-only baseline to evaluate the text-bias in the automatically curated train set.
    \item \textbf{2D-TAN \cite{2d-tan}, VSLNet \cite{vslnet}, UMT \cite{umt}:} %
    All these prior models aim to localize text in videos. %
    To apply them for detour window localization, we %
    train them to %
    accept video context as input, namely using %
    a visual mapper $f_{vm}$ similar to our approach---thus providing a late fusion of video $V_s$ context. These visual tokens are prepended to the text token, same as our token sequence (see Fig. \ref{fig:model}). This enhancement enables us to evaluate ``with $V_s$'' and ``with $V_s$, $\mathcal{Q}$'' in addition to the standard ``with $\mathcal{Q}$''

    \item \textbf{STALE \cite{stale}:} This is \emph{zero-shot} temporal detection method uses vision-language prompting. 
    Same as above, we evaluate this baseline at three combination on inputs.
\end{itemize}

The baselines in \cite{2d-tan,vslnet,umt} are trained on our same detour dataset for localization, whereas \cite{stale} is zero-shot.

\textbf{Metrics.} Following \cite{ego4d,vslnet,2d-tan}, we report recall@1 for IoU thresholds in $[0.3, 0.5, 0.7]$. We also report the recall@$1$ at the average IoU. All the recall metrics range in $[0, 1]$, higher the better.

\textbf{Results.} Tab.~\ref{tab:all-results-main} shows the results with %
greatest input context 
(``with $V_s$, $\mathcal{Q}$''), and Tab.~\ref{tab:retrieval-ablation} (right) compares against different input combinations.
We  outperform all the baselines and ablations by a clear margin. Our mean recall@1 is 3.2\% higher than the second best performing method, STALE \cite{stale}. The same trend is true for all IoU thresholds, with higher thresholds having lower recall, as expected. Again, the LLM-based parser is better for obtaining the timestamps, and using narrations is better than an external knowledge base or captions.

Our gains can be attributed to our %
model design involving early fusion of previous viewing context and the query features. The detour queries have less context than in a typical text localization task. For example, \emph{``can I skip adding salt here?''} requires a model to first understand the step being done in the source video, followed by interpreting the query to localize a \emph{similar} step without salt. Existing methods are incapable of capturing this multimodal dependency, even if we strengthen them with late fusion. We also see that our mean recall@$1$ performance is better than all other methods at all input combinations, including ablations. 

Fig.~\ref{fig:main_viz} also shows example detour localizations. %
We see that our  method is able to use the source video context and the user query to localize the detour window. For example, when a user asks \emph{``can I do this step with a steel plank saver''}, our model correctly localizes the use of a steel plank saver to put the salmon of the grill in the target video. %

\section{Conclusion}

We propose a novel task of finding detours for navigating instructional videos.  Building on video and language modeling, we develop a weakly-supervised training dataset and a novel method to train a detours network.  Our results 
show how existing %
methods are insufficient to address this problem.  %
The dataset will be released to the community to support research in navigating instructional videos.   %

\noindent\small{\noindent \textbf{Acknowledgements:} UT Austin is supported in part by the IFML NSF AI Institute. KG is paid as a research scientist by Meta. We thank Suyog Jain, Austin Miller, Honey Manglani, and Robert Kuo for help with the data collection.} %

{
    \small
    \bibliographystyle{ieeenat_fullname}
    \bibliography{main}
}

\clearpage
\setcounter{page}{1}
\setcounter{section}{0}
\setcounter{figure}{0}
\setcounter{table}{0}
\maketitlesupplementary

\section{Supplementary video}

We attach a supplementary video containing details about high-level idea, overview of the problem, detour dataset visualizations and some result visualizations.

\section{Dataset visualization}

We attach visualization pages that show the outputs from LLM for (a) summary generation and (b) weakly-supervised detour training data. Both these visualizations contain samples that are automatically generated using narrations with LLAMA 2. Most of the training samples are valid detour prompts with correct time windows; with some noise due to imperfections in narrations as a supervision and due to errors in LLAMA 2 generations. Finally, we also have a visualization that shows the samples from (c) manually collected testing data. These visualizations show the good quality manual annotations that we have collected.

\section{Detour dataset generation details}

In this section, we present additional details about the detour dataset generation (Sec. 3.2 in the main paper) process. We discuss the input prompt used to generate weakly-supervised detour annotations in $\mathcal{D}_D^{tr}$ and resulting sample outputs (including failure cases). Next, we detail the manually annotated dataset $\mathcal{D}_D^{te}$ collection details and sample visualizations.

\subsection{Weakly-supervised training set.}

The first step involves generating summaries from given narrations $N_i$. The narrations are obtained using ASR from narrated videos and we use the sentencified version, provided in \cite{tan}. We use the narrations along with the timestamp and provide the following prompt to LLAMA 2 \cite{llama2}:

\begin{myframe}
\textit{System: } Help summarize the steps of this\conditionaltext{YouTube} recipe whose narrations with timestamps are given. Timestamp is given in HH:MM:ss.

\noindent \textit{User: } Given the narrations of a \conditionaltext{YouTube} video, tell the recipe being made in this \conditionaltext{YouTube} video and list down the steps and start and end timestamps in the video. Answer in this format: `Recipe: Name of the recipe and brief detail 
 Step 1: [HH:MM:ss - HH:MM:ss] description of the step 
 Step 2: [HH:MM:ss - HH:MM:ss] description of the step 
 and so on'. Here are narrations with timestamps in HH:MM:ss format: $<$insert$>$
\end{myframe}

\begin{figure*}[t]
  \centering
  \includegraphics[width=\textwidth]{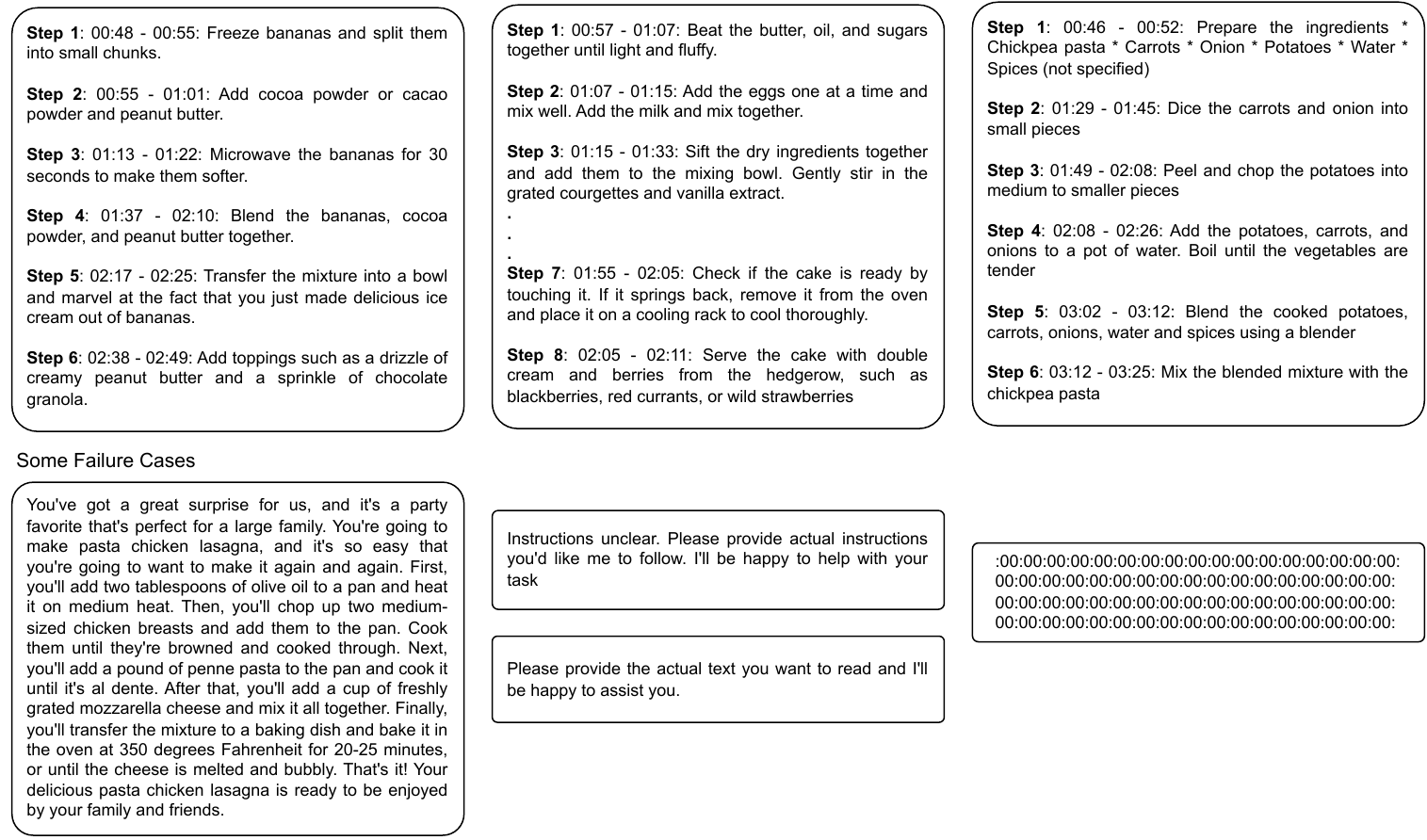}
  \caption{\textbf{Weakly-supervised summaries generated using narrations with LLAMA 2 \cite{llama2}}. While majority of the outputs contains step details and timestamps in the desired format, a few outputs are incorrect (bottom).}
  \label{fig:summary-samples}
\end{figure*}

where $<$insert$>$ is replaced by narrations with timestamps. Some sample outputs are shown in Fig.~\ref{fig:summary-samples} along with a row of failure cases (bottom). We create an automated parser that extracts steps as a tuple of timestamps and text description. Many of the videos do not contain meaningful narrations or have no narrations, and hence the outputs from these prompts do not fit into the desired output format.  They are rejected automatically by the parser. There are rare instances where even though the narrations are meaningful, the outputs are incorrect, e.g. garbage output or no output (bottom right, Fig.~\ref{fig:summary-samples}). Finally, there is a small fraction of outputs ($\sim$3\%) where the timestamps are incorrect or missing altogether (bottom left, Fig.~\ref{fig:summary-samples}). To mitigate this, we make sure steps' coverage is at least 80\% of the duration of the video. This process results in a high-quality text summary dataset. The overall process results in a summary dataset of 187K samples. Please also see the attached summary visualization page. It contains parsed summaries and we can observe the good quality summary generations using LLAMA 2 with only few failure cases.

Finally, we input two similar summaries into LLAMA 2 \cite{llama2} and generate detour instances. The process to filter similar summaries is detailed in Sec. 3.2 (main paper). For every pair of similar summaries, we use the following prompt:

\begin{myframe}
    \textit{System: } Help understand why a user would pause watching one video and take a detour to another cooking video.
    
    \noindent \textit{User: }There are two cooking videos A and B. The steps of the recipe along with timestamps in HH:MM:ss format is given. Suppose a person is watching video A, can you tell me what the user would prompt to take a detour and watch video B? The answer can be some extra/missing ingredients, tools or procedural step. Some examples of such queries can be `How to do this step without adding yeast?', `Can I add chilli powder here?',  `Can I do this step without blender?', `Can you give a video that shows other way to roll a sushi?' and so on. Also, tell the time when the user would stop watching Video A and the time range in Video B and answers the user query. Answer in this format: `Detour time in Video A: HH:MM:ss, Detour time window in Video B: [HH:MM:ss - HH:MM:ss], Detour text prompt: One sentence question a user would prompt to take a detour'. Here are the reciped: Video A: $<$insert$>$ and Video B: $<$insert$>$
\end{myframe}

\begin{figure*}[t]
  \centering
  \includegraphics[width=\textwidth]{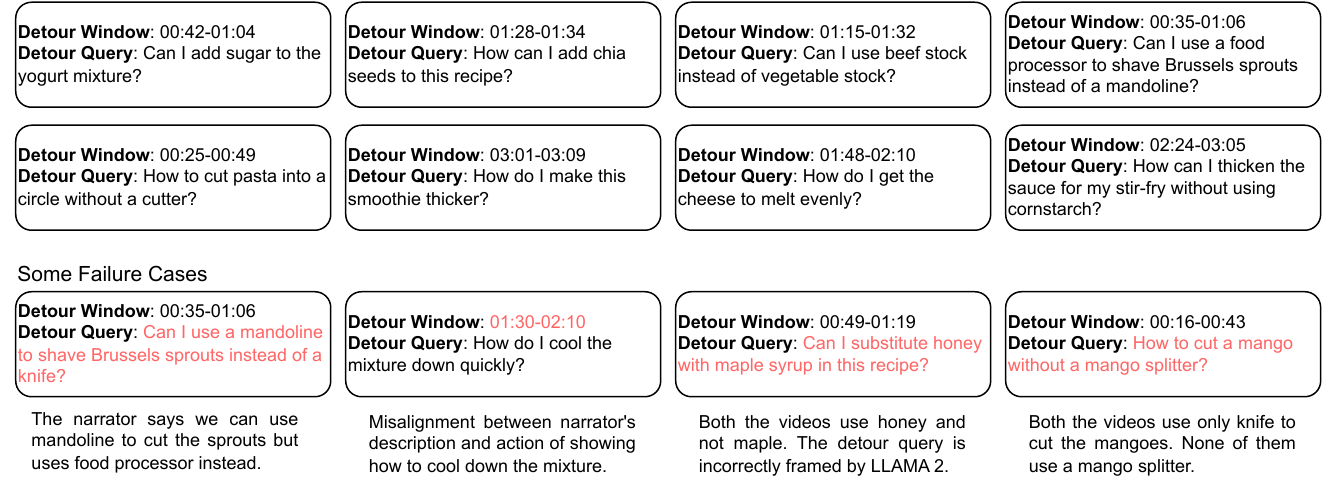}
  \caption{\textbf{Weakly-supervised detour annotation sample for training and validation.} It also contains a row of failure cases with reasons. Please also see the attached visualization for more visualizations.}
  \label{fig:detour-instances-train}
\end{figure*}

\begin{figure*}[t]
  \centering
  \includegraphics[width=0.85\textwidth]{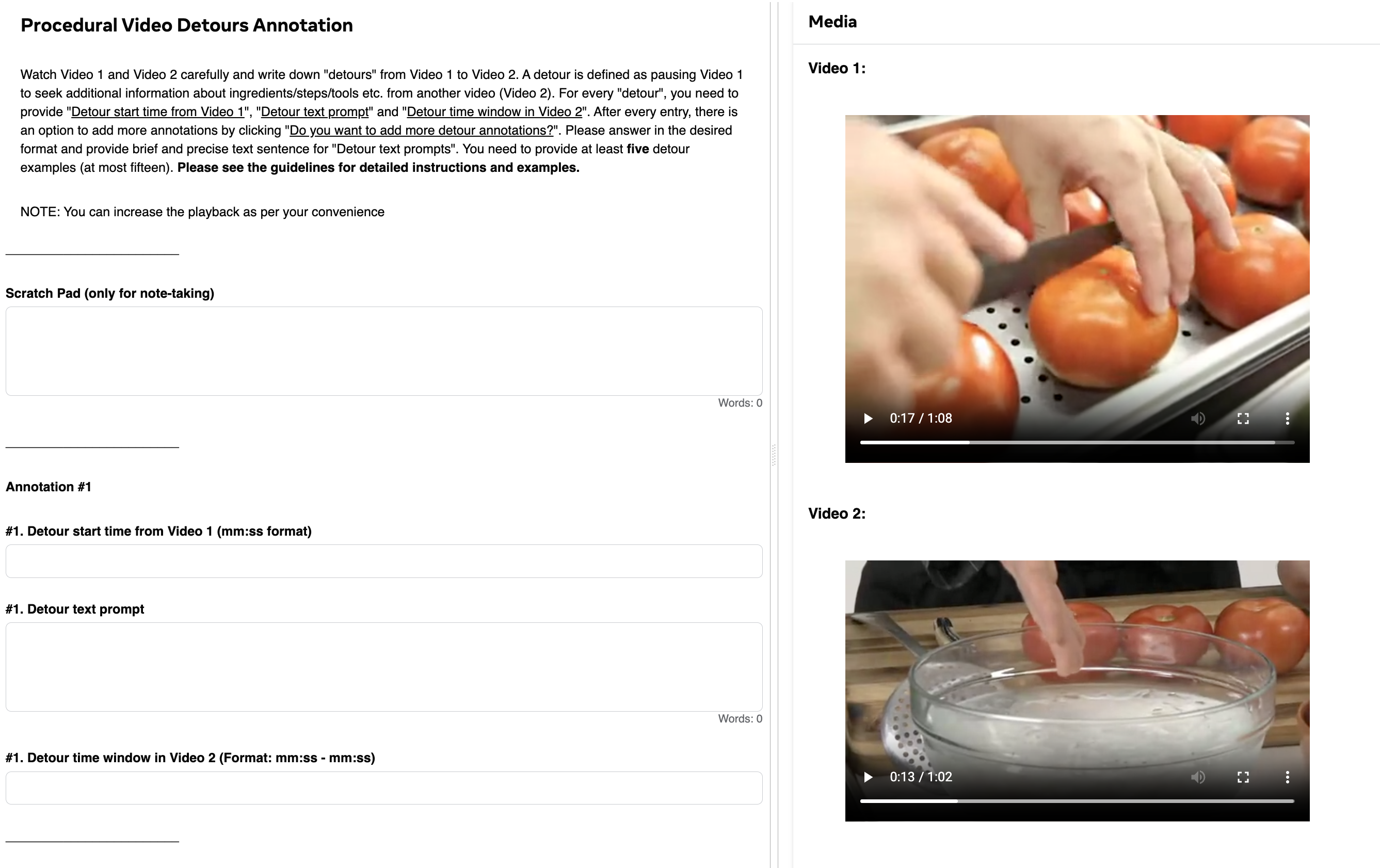}
  \caption{\textbf{Annotation interface for manual test dataset collection}. The interface reiterates important details in addition to a separate document (top). There is a scratch for the annotators to take notes while watching the two videos on the right. Finally, each instance of annotation contains a detour start time from the source video, a detour text prompt and finally a detour time window in the target video. The interface supports up to 15 annotations but only three is required.}
  \label{fig:interface}
\end{figure*}

where we insert source and detour video narrations in $<$insert$>$, respectively. Same as above, we create an automated parser to convert the text outputs into dataset tuples. A small fraction of outputs cannot be parsed by the automated parser due to incorrect output format by LLAMA 2. We ignore these instances since they are small in number in comparison to the successful parse.  Fig.~\ref{fig:detour-instances-train} shows some output samples along with failure cases. Please also see the attached visualization page for training data samples that contains automatically annotated valid detour annotations and some failure cases. We manually verify a subset of the generations and observe good quality. Furthermore, we observe a strong correlation coefficient of $>0.85$ between validation set (created automatically using narrations) and the manually collected test set across all training runs for both detour video retrieval and detour window localization task.  %

This automatically generated data is used for training only---never for ground truth evaluation of any model.

\subsection{Manually collected testing set}

We hire 24 professional annotators for manually generating video detour instances. All of them are trained for a few days on what constitutes a detour (along with examples), how to mark the time instances $t_a$ and detour window $T_d$ and what types of samples to reject. The training was followed by a pilot collection to evaluate their understanding. Finally, they annotate using a designed interface shown in Fig.~\ref{fig:interface}. We randomly sample a subset and manually verify the annotations for quality control.

\begin{figure*}[t]
  \centering
  \includegraphics[width=\textwidth]{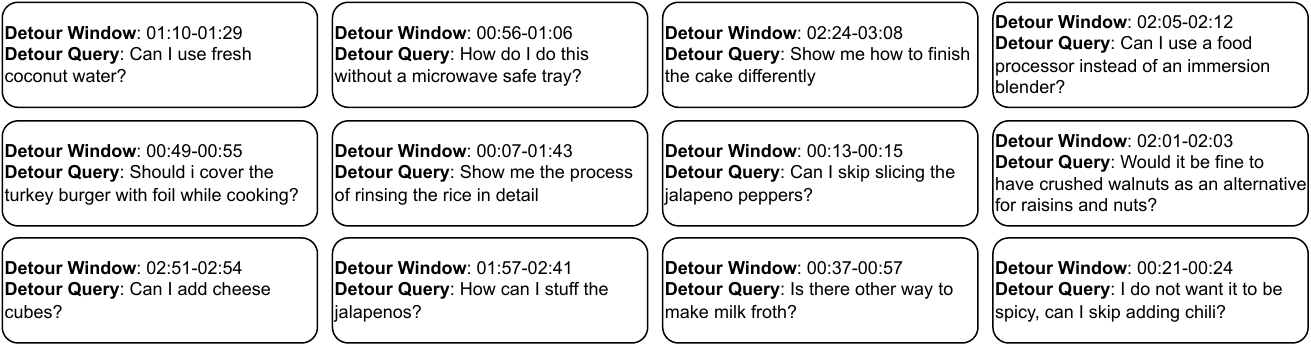}
  \caption{\textbf{Manually collected detour annotation for testing.} Please also see the attached visualization for more samples with videos that showcases the good quality annotations that we collected.}
  \label{fig:annotation-samples}
\end{figure*}

The resulting dataset consists of 3.9K source-detour video pairs resulting in $16,207$ samples. Due to our annotator trainings and quality control, the resulting dataset is of high-quality.  Fig.~\ref{fig:annotation-samples} shows representative examples. Please also see the attached visualization of test data samples that shows the high-quality samples. These manually created detours are used for evaluation.

\section{Experimental results expanded}

We expand on to the results in Sec. 4 (main paper) and show the generalizability of our method (Sec. \ref{sec:results-novel-tasks}) and performance at different input combinations (Sec. \ref{sec:results-diff-inputs}).

\subsection{Generalizability to \emph{novel} tasks}
\label{sec:results-novel-tasks}
As discussed in Sec. 3.4, we have two splits of the test data --- \emph{common} tasks containing video pairs from most frequent recipes of HowTo100M \cite{howto100m}, and \emph{novel} tasks consisting of video pairs from least frequent recipes of the dataset. We do not include any video pairs from \emph{novel} tasks in the training set to evaluate the generalizability of our method.

\textbf{Results.} Tab.  \ref{tab:supp_dataset_splits} contains the performance split for each testing subset for both detour video retrieval and detour window localization tasks. In the main paper, we showed only the performance on overall dataset for conciseness. We see that our method achieves significant gains over the baselines for both the tasks. Moreover, the performance drop in novel tasks is minimal compared to the gain. This result shows that the learned model is able to generalize to detour in newer recipes \emph{without} being explicitly trained on them. We attribute this effect to the strong interconnected nature of demonstrations in instructional videos, cooking in particular. Some recipes like \emph{making pancake} and \emph{making crepe} will only differ at some steps and detour learned with \emph{making pancake} should transfer to \emph{making crepe}.

\begin{table*}[!htb]\footnotesize
    \centering
    \setlength{\tabcolsep}{0pt} %
    \begin{tabular}{L{3.2cm}|C{1.7cm}C{1.7cm}C{1.7cm}}
    \hline
        Method & Common Task MedR $\downarrow$ & Novel Task MedR $\downarrow$ & MedR $\downarrow$   \\
        \hline
        \arrayrulecolor{lightgray} %
        \textcolor{gray}{Text-only}  & \textcolor{gray}{508} & \textcolor{gray}{524} &  \textcolor{gray}{512} \\
        CLIP \cite{clip} & 348 & 310 & 342 \\
        CLIP-Hitchhiker \cite{clip-hitchhiker}  & 339 & 305 & 336 \\
        InternVideo \cite{internvideo} & 315 & 296 & 313 \\
        DistantSup. \cite{video-distant} & 320 & 350 & 329 \\
        MLLM \cite{vid2seq} & 127 & 155 & 139 \\
        CoVR \cite{covr}  & 464 & 485 & 473 \\
        \hline
        \rowcolor{Gray}
        Ours  & \textbf{29} & \textbf{35}  & \textbf{30} \\
        \hline
        Ours w/o hard-negatives & 49 & 63 & 55 \\
        Ours w/ parser  & 76 & 88 & 81 \\
        \arrayrulecolor{black} %
        \hline
    \end{tabular}
    \hspace{1cm}
    \begin{tabular}{L{2.4cm}|C{1.7cm}C{1.7cm}C{1.2cm}}
    \hline
        Method & Common Task Mean R@1  & Novel Task Mean R@1  &Mean R@1   \\
        \hline
        \arrayrulecolor{lightgray} %
        \textcolor{gray}{Text-only} & \textcolor{gray}{4.0} & \textcolor{gray}{4.5} & \textcolor{gray}{4.2} \\
        2D-TAN \cite{2d-tan}   & 8.9 & 8.2  & 8.6 \\
        VSLNet \cite{vslnet}   & 9.2 & 9.8  & 9.4 \\
        UMT \cite{umt}   & 9.6 & 9.3  & 9.4 \\
        DistantSup. \cite{video-distant} & 8.8 & 7.9 & 8.3 \\
        MLLM \cite{vid2seq} & 9.7 & 10.8 & 10.2 \\
        STALE \cite{stale}   & 9.7 & 9.5  & 9.6 \\
        \rowcolor{Gray}
        Ours   & \textbf{13.3} & \textbf{12.3} & \textbf{12.8} \\
        \hline
        Ours w/ parser   & 12.0 & 11.3 &  11.6 \\
        \arrayrulecolor{black} %
        \hline
    \end{tabular}
    
    \caption{Results for detour video retrieval (left) and detour window localization (right) tasks on common task and novel task splits. Our method outperforms all prior methods and baselines by a significant margin even on novel tasks.}
    \label{tab:supp_dataset_splits}
    \vspace{\mysqueeze} %
\end{table*}

\begin{table*}[!htb]\footnotesize
    \centering
    \setlength{\tabcolsep}{0pt} %
    \begin{tabular}{L{2.4cm}|C{0.5cm}C{0.5cm}C{1.1cm}C{1.1cm}C{1.1cm}C{1.1cm}}
    \hline
        Method & $V_s$ & $\mathcal{Q}$ & R@5 & R@10 & R@50 & MedR $\downarrow$   \\
        \hline
        \arrayrulecolor{lightgray} %
        CLIP \cite{clip} &  \checkmark &  & 9.6 &13.2&26.9& 314 \\
        &   & \checkmark & 11.2 & 16.4&32.0&191 \\
        & \checkmark & \checkmark & 7.9 &11.8&25.2&342 \\
         \hline
        CLIP-Hitch. \cite{clip-hitchhiker}  & \checkmark & --- &--- &---&---&--- \\
        &  & \checkmark & 11.3 &17.7& 33.2& 186 \\
         & \checkmark & \checkmark & 8.4 &12.3&25.6& 336 \\
        \hline
        InternVideo \cite{internvideo} &  \checkmark &  & 11.2&17.0&31.8&150 \\
        &   & \checkmark & 13.1 & 19.2&37.2&138 \\
        & \checkmark & \checkmark & 9.7 &13.2&27.2&313 \\
        \hline
        DistantSup. \cite{video-distant} &  \checkmark &  & 4.9 &10.2&15.9&384 \\
        &   & \checkmark & 8.0 & 12.0 &  25.4 & 370 \\
        & \checkmark & \checkmark & 8.4 & 12.6 & 25.1 & 329 \\
        \hline
        MLLM \cite{vid2seq} &  \checkmark &  & 11.3 & 17.8 & 32.3 & 189 \\
        &   & \checkmark & 11.4 & 16.8 & 31.4 & 158 \\
        & \checkmark & \checkmark & 5.9 & 10.5 & 32.1 & 139 \\
        \hline
        CoVR \cite{covr}  & \checkmark &  & 4.9 & 10.1  & 15.9&388 \\
           &  & \checkmark & 4.1 & 10.0 & 15.6 &401  \\
          & \checkmark & \checkmark  &4.3&9.2&15.3&473  \\
        \hline
        Ours  & \checkmark &  & 6.1 &11.0&32.3&128  \\
        &  & \checkmark & 6.1 &10.8&32.6&116  \\
        \rowcolor{Gray}
        & \checkmark & \checkmark &  \textbf{17.6} & \textbf{27.8} & \textbf{62.4} & \textbf{30}  \\
        \arrayrulecolor{black} %
        \hline
    \end{tabular}
    \hspace{0.3cm}
        \begin{tabular}{L{2.4cm}|C{0.5cm}C{0.5cm}C{1.1cm}C{1.1cm}C{1.1cm}C{1.1cm}}
    \hline
        Method & $V_s$ & $\mathcal{Q}$ & R@1, IoU=0.3&R@1, IoU=0.5& R@1, IoU=0.7& Mean R@1   \\
        \hline
        \arrayrulecolor{lightgray} %
        2D-TAN \cite{2d-tan} & \checkmark & & 8.9 & 3.2 &0.9& 5.5 \\
        &   & \checkmark & 10.0 & 3.8 & 1.2 & 8.0 \\
        & \checkmark  & \checkmark & 10.3 &4.2&1.5& 8.6 \\
         \hline
        VSLNet \cite{vslnet}  & \checkmark  & & 9.2 & 3.1 & 1.1& 6.1 \\
        &   & \checkmark & 10.9 & 4.0 &1.5& 8.5 \\
         & \checkmark & \checkmark &11.8& 5.8 &1.7& 9.4 \\
        \hline
        UMT \cite{umt} & \checkmark & & 9.7&3.5&1.2& 6.5 \\
        &   & \checkmark & 11.2 & 5.4 &1.7& 8.7 \\
        & \checkmark  & \checkmark &12.0 &6.1&1.6&9.4 \\
        \hline
        DistantSup. \cite{video-distant} &  \checkmark &  & 9.8 & 3.7 & 1.2 & 7.6 \\
        &   & \checkmark & 10.0 & 3.8 & 1.2 & 7.9 \\
        & \checkmark & \checkmark & 10.6 & 4.0& 1.5 & 8.3 \\
        \hline
        MLLM \cite{vid2seq} &  \checkmark &  & 12.2& 6.0 & 1.5 & 9.1 \\
        &   & \checkmark & 12.3 & 6.2&1.7& 9.7 \\
        & \checkmark & \checkmark & 12.7 & 6.5 & 1.8 & 10.2 \\
        \hline
        STALE \cite{stale}   & \checkmark  &  & 10.0 & 3.8 & 1.2 &6.9 \\
           &   & \checkmark & 11.6 & 5.5 &1.5&8.8  \\
          &  \checkmark & \checkmark & 12.1&6.1&1.7&9.6  \\
        \hline
        Ours  & \checkmark & & 12.0 & 5.9  &1.5&8.9  \\
        &  & \checkmark & 14.6 & 6.9 &2.2&11.2  \\
        \rowcolor{Gray}
        & \checkmark & \checkmark  &\textbf{16.7}& \textbf{7.7}& \textbf{2.8} & \textbf{12.8}  \\
        \arrayrulecolor{black} %
        \hline
    \end{tabular}
    \caption{Comparison of our method with prior methods at different input combinations for detour video retrieval (left) and detour window localization (right) on all metrics. Our method outperforms all the prior works for all input combinations.}
    \label{tab:supp-expanded-ablation}
\end{table*}

\subsection{Results at different input combinations}
\label{sec:results-diff-inputs}
Tab. \ref{tab:supp-expanded-ablation} contains an expanded version of Tab. 2  from the main paper for all metrics. We showed performance only on one metric for brevity. We see that for both the tasks, the previous source video context and the query context is useful for the model. It is also interesting to note that for state-of-the-art methods InternVideo \cite{internvideo} and CLIP \cite{clip}, combining source video context directly with query features degrades the performance. This underscores the need for a smarter method to fuse the two contexts, as we show in our method.

\end{document}